\pdfoutput=1

\RequirePackage{fix-cm}  
\documentclass[11pt, a4paper, twocolumn, copyright, translate]{google}

\usepackage[authoryear, sort&compress, round]{natbib}

\uselogo{}

\usepackage{latexsym}
\usepackage{subcaption}

\usepackage[acronym, hyperfirst=false]{glossaries}
\glsdisablehyper

\usepackage[T1]{fontenc}

\usepackage[utf8]{inputenc}

\usepackage{inconsolata}

\usepackage{array}
\usepackage{collcell}
\usepackage{graphicx}
\usepackage{booktabs}
\usepackage{spverbatim}
\usepackage{xspace}
\usepackage[boldmath]{numprint}
\usepackage[table]{xcolor}
\usepackage{longtable}
\usepackage{float}

\npdecimalsign{.}

\newcolumntype{H}{>{\setbox0=\hbox\bgroup}c<{\egroup}@{}}

\makeatletter
\def\numprintwithbold#1{\@numprintwithbold#1\relax}
\def\@numprintwithbold#1#2\relax{%
  \if#1b%
    \textbf{\numprint{#2}}%
  \else
    \numprint{#1#2}%
  \fi
}
\makeatother

\newcolumntype{R}[1]{>{\nprounddigits{#1}\collectcell\numprintwithbold}r<{\endcollectcell}}

\usepackage{mdframed}

\title{\gemtrans Technical Report}
\author{
  Google Translate Research Team
}
\correspondingauthor{vilar@google.com. See Contributions section for full author list.}

\newcommand{\metricx}{MetricX\xspace}

\newcommand{\comet}{\textsc{Comet22}\xspace}

\newcommand{\wmtpp}{WMT24\nolinebreak\hspace{-.05em}\raisebox{.4ex}{\tiny\bf +}\nolinebreak\hspace{-.10em}\raisebox{.4ex}{\tiny\bf +}\xspace}

\defcitealias{gemma3technicalreport}{Gemma Team, 2025}
\newcommand{\citegemma}{\citepalias{gemma3technicalreport}}

\newcommand{\citepgemma}{\citepalias{gemma3technicalreport}}

\defcitealias{gemini25technicalreport}{Gemini Team, 2025}

\newcommand{\citepgemini}{\citepalias{gemini25technicalreport}}

\newcommand{\gemmathree}{Gemma~3\xspace}
\newcommand{\gemtrans}{TranslateGemma\xspace}  

\newacronym{mqm}{MQM}{Multidimensional Quality Metrics}
\newacronym{rl}{RL}{Reinforcement Learning}
\newacronym{sft}{SFT}{Supervised Fine-tuning}

\begin{abstract}
We present \gemtrans, a suite of open machine translation models based on the \gemmathree foundation models.
To enhance the inherent multilingual capabilities of \gemmathree for the translation task, we employ a two-stage fine-tuning process.
First, supervised fine-tuning is performed using a rich mixture of high-quality large-scale synthetic parallel data generated via state-of-the-art models and human-translated parallel data.
This is followed by a reinforcement learning phase, where we optimize translation quality using an ensemble of reward models, including \metricx-QE and AutoMQM, targeting translation quality.
We demonstrate the effectiveness of \gemtrans with human evaluation on the WMT25 test set across 10 language pairs and with automatic evaluation on the \wmtpp benchmark across 55 language pairs.
Automatic metrics show consistent and substantial gains over the baseline \gemmathree models across all sizes.
Notably, smaller \gemtrans models often achieve performance comparable to larger baseline models, offering improved efficiency.
We also show that \gemtrans models retain strong multimodal capabilities, with enhanced performance on the Vistra image translation benchmark.
The release of the open \gemtrans models aims to provide the research community with powerful and adaptable tools for machine translation.
\end{abstract}

\begin{document}
\maketitle

%
%

\section{Introduction}
\label{sec:introduction}

In an increasingly interconnected world, machine translation~(MT) plays a pivotal role in bridging language barriers, facilitating global communication, and democratizing access to information.
The development of large language models~(LLMs) has significantly advanced the state-of-the-art in MT.
However, progress is greatly benefitted by the availability of strong, open models that allow for transparency, reproducibility, and community-driven innovation.

To this end, we present \gemtrans, an open variant of the \gemmathree foundation model \citepgemma, specifically enhanced for machine translation.
While \gemmathree is already a potent multilingual LLM, \gemtrans has been further refined to deliver superior translation quality.
This improvement is achieved through a two-stage process: \gls{sft} on a diverse corpus of parallel data (Section~\ref{sec:sft}) and \gls{rl} from human and model-based feedback (Section~\ref{sec:rl}).

Our SFT approach leverages a blend of human-translated and synthetically-generated parallel texts, carefully curated to improve translation quality without compromising the model's general capabilities.
The RL phase employs a combination of reward models designed to optimize translation quality.
We demonstrate the efficacy of \gemtrans on the WMT25 and \wmtpp datasets, showing substantial gains across 55 language pairs.

Furthermore, \gemtrans retains the inherent multimodal capabilities of the original \gemmathree model.
Our experiments on the Vistra corpus \citep{salesky-etal-2024-benchmarking} indicate that the enhancements in text translation also positively impact image translation performance, showcasing the model's versatility.
We believe the release of \gemtrans will provide a valuable resource for researchers and practitioners in the field of machine translation.

\section{Training data}
\label{sec:training-data}

We use two types of data for the training of the models, most of it shared between the \gls{sft} and \gls{rl} phases.

\subsection{Synthetic Gemini-Generated Translation Data}

Our goal is to generate high-quality synthetic data for each language, as this has been shown to greatly improve translation quality \citep{finkelstein-etal-2024-introducing}.
As the source of monolingual data we use the MADLAD-400 corpus \citep{kudugunta2023madlad}.

We aim to produce up to 10K synthetic examples per language pair.
In order to select the source sentences that potentially benefit more from the synthetic data generation, we first bucket the original segments by length.
We then sample each bucket to obtain 1 million source segments for each language pair we wish to generate synthetic data for.
We then run a preliminary filtering step across these source segments where we take 2 samples from Gemini~2.5 Flash, once using greedy decoding and once  sampled with a temperature of~$1.0$ and compare their scores according to \metricx24-QE~\citep{juraska2024metricx}.
We select the sources where the sample achieves the largest improvement over the greedy decoding.
The intuition behind this source filtering approach is that we wish to select sources that will benefit the most from 128-sample QE decoding, so we use 2 samples as a low-cost approximation.

After this selection process, for each of the sources for each language pair we generate 128 samples from Gemini~2.5 Flash and then apply a \metricx24-QE filter to select the best-performing examples.
We generate translations of two distinct lengths this way: individual sentences and text blobs of up to 512 tokens.
This way we aim to support both translations of individual segments as well as longer texts.
For generating these translations we used the same prompt as we used for further training (see Section~\ref{sec:prompting}).
In order to avoid formatting issues or erroneous translations, we apply an additional formatting filtering step, again based on Gemini~2.5 Flash.
This methodology was applied for all language pairs that are covered by \wmtpp~\citep{deutsch2025wmt24++} plus an additional set of 30 language pairs that are specified in Appendix~\ref{sec:synth-data-langs}.

\subsection{Human-Generated Translation Data}
To increase the diversity and script coverage of the data we also include data for additional lower-resource languages.
For these languages, we opt to use human-generated parallel data instead.
This data comes from the SMOL~\citep{caswell2025smol} and GATITOS~\citep{jones-etal-2023-gatitos} datasets.
SMOL covers 123 languages and GATITOS covers 170.

\subsection{Language distribution}

\begin{figure*}[ht]
    \centering
    \begin{subfigure}[b]{0.48\textwidth}
       \centering
       \includegraphics[width=\textwidth]{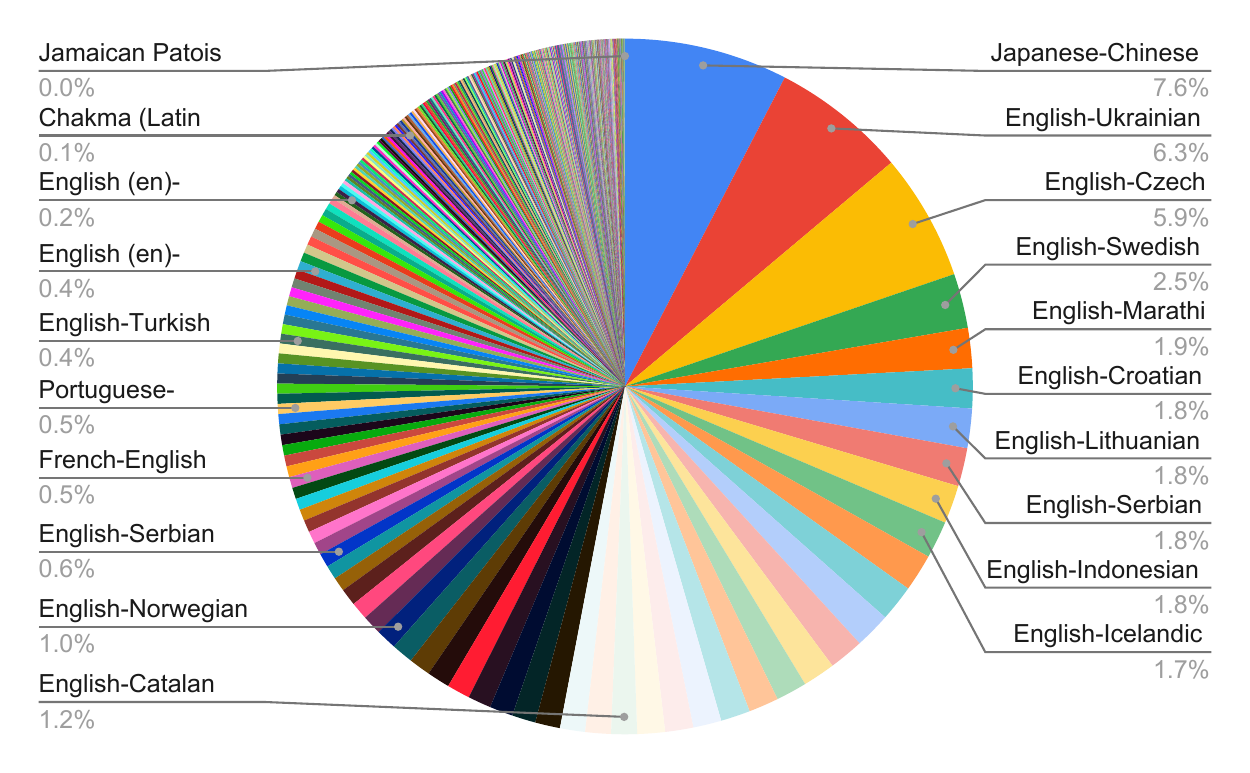}
       \caption{SFT data mixture.}
       \label{fig:subfig_left}
    \end{subfigure}
    \hfill
    \begin{subfigure}[b]{0.48\textwidth}
       \centering
       \includegraphics[width=\textwidth]{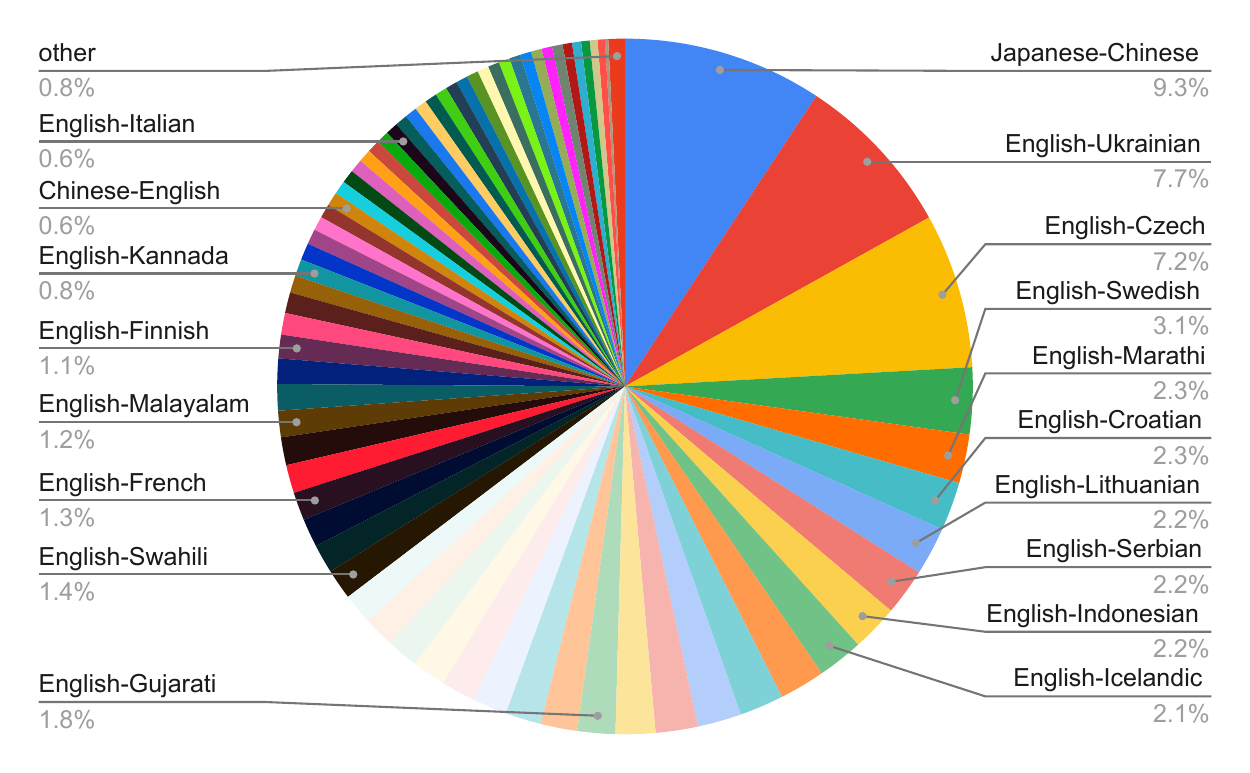}
       \caption{RL data mixture.}
       \label{fig:subfig_right}
    \end{subfigure}

    \caption{Language distribution in the \gemtrans data mixtures measured as model tokens.}
    \label{fig:data_mixture}
\end{figure*}
The final proportion of languages for the SFT and RL phases can be found in Figure~\ref{fig:data_mixture}.
For \gls{rl} we used the same translation data as for SFT, except for GATITOS and SMOL that were used in SFT only.
We provide the full list of languages that were included in training in Appendix~\ref{sec:list-of-languages}.

\subsection{Generic Instruction-Following Data}

Our SFT mixture also includes 30\% generic instruction-following data from the original \gemmathree mixture.
The purpose of including this data is to prevent the model from overfitting to the translation task and to maintain generic instruction-following capabilities.

\section{Supervised Fine-Tuning}
\label{sec:sft}

For supervised fine-tuning (SFT), we begin with the released \gemmathree~27B, 12B and 4B checkpoints.
We use parallel data including both human-generated texts as well as synthetic data generated by Gemini \citepgemini, as described in Section~\ref{sec:training-data}.
In addition we use generic instruction-following data.
We use the Kauldron SFT tooling\footnote{\url{https://kauldron.readthedocs.io/en/latest/}} to fine-tune the \gemmathree checkpoints.
For fine-tuning we use the AdaFactor optimizer \citep{shazeer2018adafactor} with a learning rate of 0.0001 and a batch size of 64, running for 200k steps.
We update all model parameters, but freeze the embedding parameters, as preliminary experiments indicated this helped with translation performance for languages and scripts not covered in the SFT data mix.

\section{Reinforcement Learning}
\label{sec:rl}

We performed reinforcement learning on top of the SFT checkpoint, using an ensemble of metrics as reward models, to further boost translation quality.

We used the following metrics as reward models during RL:
\begin{itemize}
    \item MetricX-24-XXL-QE \citep{juraska2024metricx}, a learned, regression-based translation metric producing a floating point score between 0 (best) and 25 (worst), matching the standard \gls{mqm} score range \citep{freitag2021experts}.
\metricx scores were linearly rescaled, using $5.0 - \texttt{score}$, when computing rewards, so that higher scores indicate better quality. Although MetricX can take source, reference, and hypothesis as input, we used it as a QE metric by passing in an empty reference.

    \item Gemma-AutoMQM-QE, a finetuned AutoMQM model~\citep{fernandes2023devil}.
This model was initialized from the \gemmathree-27B-IT checkpoint \citepgemma, and was trained on \gls{mqm} ratings data from WMT 2020 - WMT 2023 \citep{lommel-etal-mqm,freitag2021experts}.
Default \gls{mqm} weights \citep{freitag2021experts} were used in computing (token-level) rewards from AutoMQM outputs. As with MetricX, it ignores the reference translation.

    \item ChrF \citep{popovic2015chrf}, a lexical overlap-based translation metric. This was the only reward model for which the (synthetic) references were used. ChrF scores were scaled by a factor of two to be on approximately the same scale as the other rewards.

    \item Naturalness Autorater developed in-house, using the base RL policy model as a prompted LLM-as-a-Judge. As with AutoMQM, this Autorater elicited span-level annotations. This Autorater was instructed to penalize spans in the machine-translated text which did not sound like they were produced by a native speaker (conditioned on the naturalness errors in the output \textit{not} stemming from an unnatural source input).

    \item Generalist reward model covering many tasks, including reasoning, instruction following, and multilingual abilities, adapted from the general \gemmathree post-training setup \citegemma.

\end{itemize}

\begin{figure*}
\centering
\includegraphics[width=0.8\textwidth]{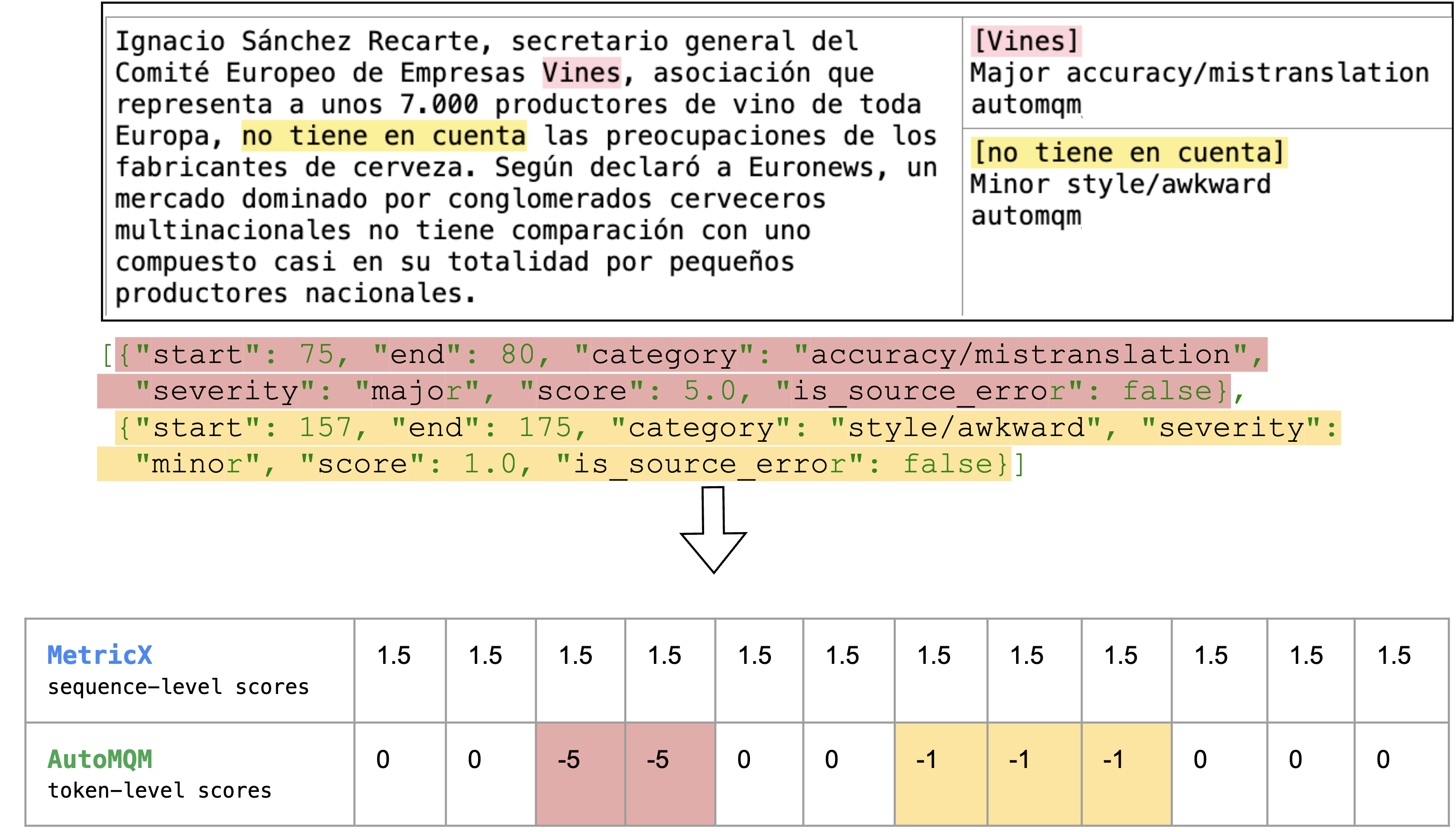}
\caption{Illustration of how sequence-level and token-level rewards are additively combined during advantage computation in RL. Note that advantage is computed from sequence-level rewards as `reward-to-go', meaning that rewards are broadcast uniformly to every token.}
\label{fig:token_rl}
\end{figure*}

We used \gls{rl} algorithms extended to support token-level advantages, which were added to the advantages computed from sequence-level rewards.
This allowed us to use fine-grained, span-level reward signals from AutoMQM and the Naturalness Autorater directly, for improved credit assignment and training efficiency in the spirit of \citet{ramos2024fine}. See Figure \ref{fig:token_rl} for an illustration of how \metricx and AutoMQM rewards were (additively) combined during advantage computation.
The combined advantages were then batch-normalized.

\section{Automatic Evaluation}
\label{sec:automatic-eval}

\subsection{Text translation}
\label{sec:text-translation}

We evaluate \gemtrans using \metricx24 \citep{juraska2024metricx} and \comet \citep{rei2022comet}.
The \gemtrans models consistently show improved translation quality compared to the baseline \gemmathree models across all evaluated sizes and metrics, as detailed in Table~\ref{tab:automatic-results}.

\begin{table}
  \caption{Automatic evaluation results using \metricx and \comet~(C22) on \wmtpp.}
  \label{tab:automatic-results}
  \begin{center}
    \begin{tabular}{llR{2}R{1}H}
      \toprule
      Size
      & System
      & \multicolumn{1}{r}{\makebox[0em][r]{MetricX$\downarrow$}}
      & \multicolumn{1}{r}{C22$\uparrow$}
      & \\ 
      \midrule
      27B & \gemmathree & 4.037  & 83.1  & 50.5 \\
          & \gemtrans   & b3.087 & b84.4 & 49.5 \\
      \midrule
      12B & \gemmathree & 4.863  & 81.6  & 47.7 \\
          & \gemtrans   & b3.601 & b83.5 & 47.5 \\
      \midrule
      4B  & \gemmathree & 6.967  & 77.2  & 44.0 \\
          & \gemtrans   & b5.323 & b80.1 & 43.5 \\
      \bottomrule
    \end{tabular}
  \end{center}
\end{table}

For the 27B parameter model, the \gemtrans version attains an average MetricX score of 3.09, a substantial reduction from the baseline \gemmathree's score of 4.04.
This represents a relative decrease of approximately 23.5\%, signaling a marked increase in translation fidelity.
Similar trends are observed for the other model sizes.
The 12B \gemtrans model achieves a MetricX of 3.60, down from 4.86 for the baseline (a 25.9\% reduction), while the 4B \gemtrans model scores 5.32, compared to 6.97 for the baseline (a 23.6\% reduction).

\comet confirms the trend of improvements for the \gemtrans model.
In addition, this shows that improvements carry over to metrics not explicitly optimized for in the RL phase.
For instance, the 12B \gemtrans model shows a score of 83.5, up from 81.6.
The 4B \gemtrans model exhibits even larger increases, with \comet rising from 77.2 to 80.1.

The effect of model scale on performance is also apparent.
As expected, larger models tend to produce better translations within both the baseline and \gemtrans series.
However, the enhancements brought by the \gemtrans fine-tuning are such that smaller \gemtrans models can achieve performance levels comparable to or even exceeding those of larger baseline models.
Notably, the 12B \gemtrans model surpasses the performance of the larger 27B baseline \gemmathree model.
Similarly, the 4B \gemtrans model achieves comparable results to the 12B baseline \gemmathree model.
This efficiency gain allows for high-quality translation with reduced computational resources.

A more granular analysis of MetricX scores for each of the 55 language pairs, presented in Appendix~\ref{sec:results-per-lang}, reveals that the improvements of \gemtrans are consistent across all 55 language pairs evaluated.
Some example improvements for specific languages are
\begin{itemize}
    \item English to German: 1.63 down to 1.19,
    \item English to Spanish: 2.54 down to 1.88,
    \item English to Hebrew: 3.90 down to 2.72,
    \item English to Swahili: 5.92 down to 4.45,
    \item English to Lithuanian: 6.01 down to 4.39,
    \item English to Estonian: 6.40 down to 4.61 and
    \item English to Icelandic: 8.31 down to 5.69.
\end{itemize}
These examples highlight the model's improved ability to handle a diverse range of languages, both for high-resource languages (e.g. German, English) as well as low-resource ones (e.g. Icelandic, Swahili).

We also hypothesize that the 27B model, with its higher capacity, will have benefited more from the vast amount of languages seen during the SFT phase (detailed in Appendix~\ref{sec:list-of-languages}), although we do not have direct experimental confirmation of this.

\subsection{Prompting the Model}
\label{sec:prompting}

The model has been trained using the prompt shown in Figure~\ref{fig:preferred-prompt}, which is also the prompt we used in our evaluations.
We recommend using the same prompt for producing new translations.
Tools for automatically wrapping the text with it are provided in the model repository.

\begin{figure*}
  \begin{center}
  \begin{tabular}{>{\raggedright\arraybackslash}p{0.95\textwidth}}
    \toprule
  \texttt{You are a professional \{source\_lang\} (\{src\_lang\_code\}) to \{target\_lang\} (\{tgt\_lang\_code\}) translator. Your goal is to accurately convey the meaning and nuances of the original \{source\_lang\} text while adhering to \{target\_lang\} grammar, vocabulary, and cultural sensitivities. Produce only the \{target\_lang\} translation, without any additional explanations or commentary. Please translate the following \{source\_lang\} text into \{target\_lang\}:$\backslash$n$\backslash$n$\backslash$n\{text\}}
  \\
  \bottomrule
  \end{tabular}
  \end{center}

  \caption{Preferred prompt when using the model. \texttt{source\_lang} refers to the source language name, e.g. English, \texttt{src\_lang\_code} to the source language code, e.g. en-US, \texttt{target\_lang} to the target language, e.g. German, and \texttt{tgt\_lang\_code} to the target language code, i.e. de-DE.}
  \label{fig:preferred-prompt}
\end{figure*}

\subsection{Image Translation}
\label{sec:image-translation}
\begin{figure}
\centering
   \centering
   \includegraphics[width=0.48\textwidth]{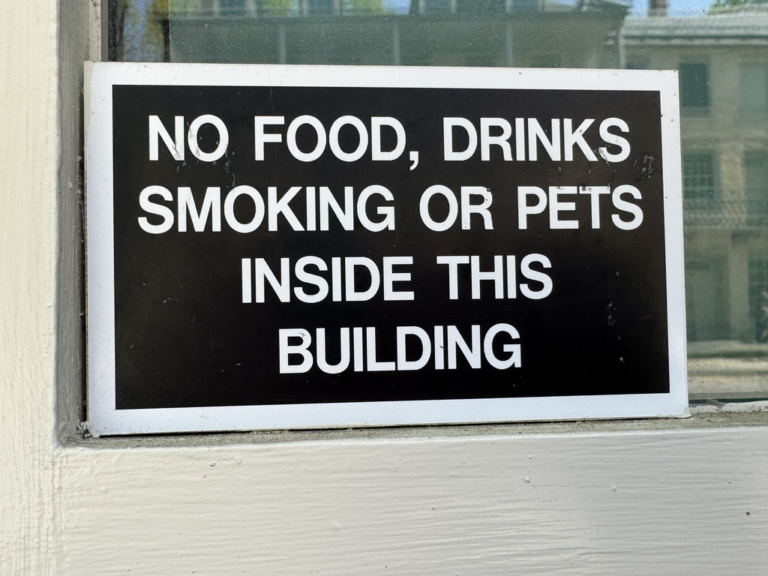}
   \caption{Example of the pictures included in the Vistra benchmark.}
   \label{fig:no_food_inside}
\end{figure}

We used the Vistra benchmark \citep{salesky-etal-2024-benchmarking} to assess whether the models retained their ability to translate text within images after our additional training steps.
Note that no multimodal training data was used in the \gls{sft} or \gls{rl} steps reported in this work.
In order to simplify the evaluation protocol, we selected only images that, according to the reference, contained a single instance of text.
This resulted in a set of 264 images.
An example is shown in Figure~\ref{fig:no_food_inside}.
The input to the model was just the image together with a prompt asking it to translate the text in it.\footnote{The model release also includes an interface for image translation, similar to the one for text translation.}
In particular, we did not include any other information about the text, like its location in the image or a previous OCR pass.

The results, presented in Table~\ref{tab:automatic-image-translation-results}, show that \gemtrans retains the image processing capabilities of the base \gemmathree models.
The improvements in translation quality attained by \gemtrans carry over for this task, with the exception of the 12B model measured in \comet.
We see MetricX score improvements of nearly 0.5 points in the case of the 27B model, or 0.25 for the 12B model.

The smaller 4B model obtains only small improvements when compared to the baseline, probably due to its limited capacity.

\begin{table}
  \caption{
  Automatic evaluation results using \metricx and \comet~(C22) for image translation performance, on the Vistra corpus.
  The scores are the average of translating from English into German, Spanish, Russian and Chinese.
}
  \label{tab:automatic-image-translation-results}
  \begin{center}
    \begin{tabular}{llR{2}R{1}}
      \toprule
      Size
      & System
      & \multicolumn{1}{r}{\makebox[0em][r]{MetricX$\downarrow$}}
      & \multicolumn{1}{r}{C22$\uparrow$} \\
      \midrule
      27B & \gemmathree & 2.025  & 76.075  \\
          & \gemtrans   & b1.575 & b77.65  \\
      \midrule
      12B & \gemmathree & 2.325  & b74.925 \\
          & \gemtrans   & b2.075 & 72.775  \\
      \midrule
      4B  & \gemmathree & 2.600  & 69.125  \\
          & \gemtrans   & b2.575 & b70.725 \\
      \bottomrule
    \end{tabular}
  \end{center}
\end{table}

\section{Human Evaluation}
\label{sec:human-eval}
We conduct an additional human evaluation on a limited set of language directions to measure \gemtrans's translation performance.
We do so using MQM \citep{lommel-etal-mqm, freitag2021experts}, a human evaluation framework where professional translators highlight error spans in translations, with document context, assigning a severity and category to each, with a score being automatically derived by counting the errors with a weighting scheme.
We collected the annotations using the open-source Anthea tool.\footnote{\url{https://github.com/google-research/google-research/tree/master/anthea}}
We evaluate the models in 10 language pairs, from 3 distinct source languages:
\begin{itemize}
  \item English to German
  \item English to Chinese (Simplified)
  \item English to Italian
  \item English to Serbian (Cyrillic)
  \item English to Korean
  \item English to Swahili (Kenyan)
  \item English to Marathi
  \item Czech to Ukrainian
  \item Czech to German
  \item Japanese to English
\end{itemize}
We selected this set to have a mix of high- and low-resource languages, in addition to having different language families and writing systems.
The source data is all taken from the WMT25 translation task, using the literary, news, and social domains.
For all language pairs, we evaluated \gemtrans~12B and 27B, as well as \gemmathree 27B.

To avoid issues with rater fatigue, each document in the dataset was truncated at paragraph boundaries to have no more than 12 source sentences, skipping documents with more than 12 sentences in the first paragraph.
However, for the literary domain, where each document is an entire book chapter, documents were split into ``chunks'' of 1 or more paragraphs up to the 12-sentence limit, with each chunk being human-evaluated in isolation.
Following \citet{riley-etal-2024-finding}, we used a ``pseudo-SxS'' rater assignment, where all system outputs for a particular source document were evaluated by the same rater.

\begin{table*}
  \caption{MQM results of the human evaluation for \gemtrans and \gemmathree. Lower scores are better.}
  \label{tab:human-evaluation}
  \begin{center}
    \begin{tabular}{lR{1}R{1}R{1}}
      \toprule
                    & \multicolumn{2}{c}{\gemtrans}
                    & \multicolumn{1}{c}{\gemmathree} \\
                    \cmidrule{2-3}
      Language Pair &
      \multicolumn{1}{r}{27B} &
      \multicolumn{1}{r}{12B} &
      \multicolumn{1}{r}{27B} \\
      \midrule
      English$\to$Italian  & b1.813  & 2.035  & 2.453   \\
      English$\to$German   & b2.252  & 3.177  & b2.158  \\
      English$\to$Marathi  & b3.082  & 4.647  & 4.657   \\
      English$\to$Korean   & b3.103  & 4.57   & 3.785   \\
      English$\to$Swahili  & b4.229  & 5.191  & 5.216   \\
      Czech$\to$Ukrainian  & b5.323  & 8.49   & 6.294   \\
      English$\to$Chinese  & b6.343  & 8.432  & 7.415   \\
      English$\to$Serbian  & b8.737  & 15.759 & 10.381  \\
      Czech$\to$German     & b10.311 & 11.371 & b10.240 \\
      Japanese$\to$English & 13.391  & 15.739 & b11.565 \\
      \bottomrule
    \end{tabular}
  \end{center}
\end{table*}

The results can be found in Table~\ref{tab:human-evaluation}.
For most language pairs, the human evaluation confirms the trend we see on the automatic metrics, with \gemtrans clearly outperforming \gemmathree.
There are two exceptions: when the target language is German, where both models are on par, and Japanese$\to$English where \gemtrans actually suffers a regression.
Looking into the error categorization, we found that this is due to mistranslation of named entities, while other error categories did improve.

The improvements for \gemtrans are especially relevant for low-resource language pairs.
E.g. for English to Marathi we obtain an improvement of 1.6 points, or 1.0 for English to Swahili or Czech to Ukrainian.
The human evaluation also confirms the performance difference between the 27B and 12B \gemtrans models already demonstrated by the automatic metrics.
That said, the 12B model still stays competitive with the bigger \gemmathree model, especially for high-resource languages.

\section{Conclusions}
\label{sec:conclusions}

In this work, we introduced \gemtrans, a series of open models based on \gemmathree, specifically enhanced for machine translation.
Through a combination of supervised fine-tuning on diverse, high-quality parallel data—blending human and synthetic sources—and a novel reinforcement learning approach utilizing an ensemble of reward models, we have improved translation performance across a wide spectrum of languages and model sizes (4B, 12B, and 27B parameters).

Our automatic evaluations on the \wmtpp dataset, encompassing 55 language pairs, show consistent gains for \gemtrans models over the baseline \gemmathree models both in \metricx and \comet.
We observed a strong performance increase across various language types, including high-resource languages like German and Spanish, and lower-resource languages such as Icelandic or Swahili.
A key finding is the enhanced efficiency of the \gemtrans models, where smaller fine-tuned models often match or exceed the performance of larger baseline models, offering a better trade-off between quality and computational cost.

Furthermore, we have shown that \gemtrans models retain the multimodal capabilities of the original \gemmathree.
Experiments on the Vistra benchmark indicate that the improvements in text translation extend to the image translation task, particularly for the 12B and 27B models, without any specific multimodal fine-tuning.

The release of the \gemtrans models contributes a valuable set of open-source tools for the machine translation community, fostering further research and application development.
We believe these models will serve as a strong foundation for a variety of translation-related tasks and encourage their adoption and exploration.

\bibliography{custom}

@misc{gemma3technicalreport,
      title={Gemma 3 Technical Report},
      author={{Gemma Team}
              and Aishwarya Kamath
              and Johan Ferret
              and Shreya Pathak
              and Nino Vieillard
              and Ramona Merhej
              and Sarah Perrin
              and Tatiana Matejovicova
              and Alexandre Ramé
              and Morgane Rivière
              and Louis Rouillard
              and others},
  year={2025},
  eprint={2503.19786},
  archivePrefix={arXiv},
  primaryClass={cs.CL},
  url={https://arxiv.org/abs/2503.19786},
}

@misc{gemini25technicalreport,
      title={Gemini 2.5: Pushing the Frontier with Advanced Reasoning, Multimodality, Long Context, and Next Generation Agentic Capabilities},
      author={{Gemini Team}
        and Gheorghe Comanici
        and Eric Bieber
        and Mike Schaekermann
        and Ice Pasupat
        and Noveen Sachdeva
        and Inderjit Dhillon
        and Marcel Blistein
        and Ori Ram
        and Dan Zhang
        and Evan Rosen
        and others},
      year={2025},
      eprint={2507.06261},
      archivePrefix={arXiv},
      primaryClass={cs.CL},
      url={https://arxiv.org/abs/2507.06261},
}

@inproceedings{juraska2024metricx,
    title = "{M}etric{X}-24: The {G}oogle Submission to the {WMT} 2024 Metrics Shared Task",
    author = "Juraska, Juraj  and
      Deutsch, Daniel  and
      Finkelstein, Mara  and
      Freitag, Markus",
    editor = "Haddow, Barry  and
      Kocmi, Tom  and
      Koehn, Philipp  and
      Monz, Christof",
    booktitle = "Proceedings of the Ninth Conference on Machine Translation",
    month = nov,
    year = "2024",
    address = "Miami, Florida, USA",
    publisher = "Association for Computational Linguistics",
    url = "https://aclanthology.org/2024.wmt-1.35/",
    doi = "10.18653/v1/2024.wmt-1.35",
    pages = "492--504"
}

@inproceedings{fernandes2023devil,
  title = "The Devil Is in the Errors: Leveraging Large Language Models for Fine-grained Machine Translation Evaluation",
  author = "Fernandes, Patrick  and
  Deutsch, Daniel  and
  Finkelstein, Mara  and
  Riley, Parker  and
  Martins, Andr{\'e}  and
  Neubig, Graham  and
  Garg, Ankush  and
  Clark, Jonathan  and
  Freitag, Markus  and
  Firat, Orhan",
  editor = "Koehn, Philipp  and
  Haddow, Barry  and
  Kocmi, Tom  and
  Monz, Christof",
  booktitle = "Proceedings of the Eighth Conference on Machine Translation",
  month = dec,
  year = "2023",
  address = "Singapore",
  publisher = "Association for Computational Linguistics",
  url = "https://aclanthology.org/2023.wmt-1.100/",
  doi = "10.18653/v1/2023.wmt-1.100",
  pages = "1066--1083"
}

@article{freitag2021experts,
  title = "Experts, Errors, and Context: A Large-Scale Study of Human Evaluation for Machine Translation",
  author = "Freitag, Markus  and
  Foster, George  and
  Grangier, David  and
  Ratnakar, Viresh  and
  Tan, Qijun  and
  Macherey, Wolfgang",
  editor = "Roark, Brian  and
  Nenkova, Ani",
  journal = "Transactions of the Association for Computational Linguistics",
  volume = "9",
  year = "2021",
  address = "Cambridge, MA",
  publisher = "MIT Press",
  url = "https://aclanthology.org/2021.tacl-1.87/",
  doi = "10.1162/tacl_a_00437",
  pages = "1460--1474"
}

@inproceedings{rei2022comet,
  title = "{COMET}-22: Unbabel-{IST} 2022 Submission for the Metrics Shared Task",
  author = "Rei, Ricardo  and
  C. de Souza, Jos{\'e} G.  and
  Alves, Duarte  and
  Zerva, Chrysoula  and
  Farinha, Ana C  and
  Glushkova, Taisiya  and
  Lavie, Alon  and
  Coheur, Luisa  and
  Martins, Andr{\'e} F. T.",
  editor = {Koehn, Philipp  and
            Barrault, Lo{\"i}c  and
            Bojar, Ond{\v{r}}ej  and
            Bougares, Fethi  and
            Chatterjee, Rajen  and
            Costa-juss{\`a}, Marta R.  and
            Federmann, Christian  and
            Fishel, Mark  and
            Fraser, Alexander  and
            Freitag, Markus  and
            Graham, Yvette  and
            Grundkiewicz, Roman  and
            Guzman, Paco  and
            Haddow, Barry  and
            Huck, Matthias  and
            Jimeno Yepes, Antonio  and
            Kocmi, Tom  and
            Martins, Andr{\'e}  and
            Morishita, Makoto  and
            Monz, Christof  and
            Nagata, Masaaki  and
            Nakazawa, Toshiaki  and
            Negri, Matteo  and
            N{\'e}v{\'e}ol, Aur{\'e}lie  and
            Neves, Mariana  and
            Popel, Martin  and
            Turchi, Marco  and
            Zampieri, Marcos},
  booktitle = "Proceedings of the Seventh Conference on Machine Translation (WMT)",
  month = dec,
  year = "2022",
  address = "Abu Dhabi, United Arab Emirates (Hybrid)",
  publisher = "Association for Computational Linguistics",
  url = "https://aclanthology.org/2022.wmt-1.52/",
  pages = "578--585"
}

@inproceedings{deutsch2025wmt24++,
  title = "{WMT}24++: Expanding the Language Coverage of {WMT}24 to 55 Languages {\&} Dialects",
  author = "Deutsch, Daniel  and
  Briakou, Eleftheria  and
  Caswell, Isaac Rayburn  and
  Finkelstein, Mara  and
  Galor, Rebecca  and
  Juraska, Juraj  and
  Kovacs, Geza  and
  Lui, Alison  and
  Rei, Ricardo  and
  Riesa, Jason  and
  Rijhwani, Shruti  and
  Riley, Parker  and
  Salesky, Elizabeth  and
  Trabelsi, Firas  and
  Winkler, Stephanie  and
  Zhang, Biao  and
  Freitag, Markus",
  editor = "Che, Wanxiang  and
  Nabende, Joyce  and
  Shutova, Ekaterina  and
  Pilehvar, Mohammad Taher",
  booktitle = "Findings of the Association for Computational Linguistics: ACL 2025",
  month = jul,
  year = "2025",
  address = "Vienna, Austria",
  publisher = "Association for Computational Linguistics",
  url = "https://aclanthology.org/2025.findings-acl.634/",
  doi = "10.18653/v1/2025.findings-acl.634",
  pages = "12257--12284",
  ISBN = "979-8-89176-256-5"
}

@misc{ramos2024fine,
  title={Fine-Grained Reward Optimization for Machine Translation using Error Severity Mappings},
  author={Miguel Moura Ramos and Tomás Almeida and Daniel Vareta and Filipe Azevedo and Sweta Agrawal and Patrick Fernandes and André F. T. Martins},
  year={2025},
  eprint={2411.05986},
  archivePrefix={arXiv},
  primaryClass={cs.CL},
  url={https://arxiv.org/abs/2411.05986},
}

@article{kudugunta2023madlad,
  title={Madlad-400: A multilingual and document-level large audited dataset},
  author={Kudugunta, Sneha and Caswell, Isaac and Zhang, Biao and Garcia, Xavier and Xin, Derrick and Kusupati, Aditya and Stella, Romi and Bapna, Ankur and Firat, Orhan},
  journal={Advances in Neural Information Processing Systems},
  volume={36},
  pages={67284--67296},
  year={2023},
  url={https://arxiv.org/pdf/2309.04662},
}

@inproceedings{caswell2025smol,
  title = "{SMOL}: Professionally Translated Parallel Data for 115 Under-represented Languages",
  author = "Caswell, Isaac  and
  Nielsen, Elizabeth  and
  Luo, Jiaming  and
  Cherry, Colin  and
  Kovacs, Geza  and
  Shemtov, Hadar  and
  Talukdar, Partha  and
  Tewari, Dinesh  and
  Diane, Baba Mamadi  and
  Diane, Djibrila  and
  Ciss{\'e}, Solo Farabado  and
  Doumbouya, Koulako Moussa  and
  Ferrante, Edoardo  and
  Guasoni, Alessandro  and
  Homan, Christopher  and
  Keita, Mamadou K.  and
  DebBarma, Sudhamoy  and
  Kuzhuget, Ali  and
  Anugraha, David  and
  Shulthan Habibi, Muhammad Ravi  and
  Ahmadi, Sina  and
  Munthali, Anthony  and
  Liu, Jonathan Mingfei  and
  Eng, Jonathan",
  editor = "Haddow, Barry  and
  Kocmi, Tom  and
  Koehn, Philipp  and
  Monz, Christof",
  booktitle = "Proceedings of the Tenth Conference on Machine Translation",
  month = nov,
  year = "2025",
  address = "Suzhou, China",
  publisher = "Association for Computational Linguistics",
  url = "https://aclanthology.org/2025.wmt-1.85/",
  doi = "10.18653/v1/2025.wmt-1.85",
  pages = "1103--1123",
  ISBN = "979-8-89176-341-8"
}

@inproceedings{shazeer2018adafactor,
  title={Adafactor: Adaptive learning rates with sublinear memory cost},
  author={Shazeer, Noam and Stern, Mitchell},
  booktitle={International Conference on Machine Learning},
  pages={4596--4604},
  year={2018},
  organization={PMLR},
  url={https://proceedings.mlr.press/v80/shazeer18a/shazeer18a.pdf}
}

@inproceedings{popovic2015chrf,
  title = "chr{F}: character n-gram {F}-score for automatic {MT} evaluation",
  author = "Popovi{\'c}, Maja",
  editor = "Bojar, Ond{\v{r}}ej  and
  Chatterjee, Rajan  and
  Federmann, Christian  and
  Haddow, Barry  and
  Hokamp, Chris  and
  Huck, Matthias  and
  Logacheva, Varvara  and
  Pecina, Pavel",
  booktitle = "Proceedings of the Tenth Workshop on Statistical Machine Translation",
  month = sep,
  year = "2015",
  address = "Lisbon, Portugal",
  publisher = "Association for Computational Linguistics",
  url = "https://aclanthology.org/W15-3049/",
  doi = "10.18653/v1/W15-3049",
  pages = "392--395"
}

@article{lommel-etal-mqm,
    author = { Lommel, Arle Richard and  Burchardt, Aljoscha and  Uszkoreit, Hans },
    editor = { Görög, Attila and  Sánchez-Gijón, Pilar },
    title = {Multidimensional Quality Metrics (MQM): A Framework for Declaring and Describing Translation Quality Metrics},
    year = {2014},
    month = {12},
    number = {12},
    pages = {455--463},
    journal = {Tradumàtica: tecnologies de la traducció},
    url = {https://ddd.uab.cat/pub/tradumatica/tradumatica_a2014n12/tradumatica_a2014n12p455.pdf}
}

@inproceedings{jones-etal-2023-gatitos,
    title = "{GATITOS}: Using a New Multilingual Lexicon for Low-resource Machine Translation",
    author = "Jones, Alexander  and
      Caswell, Isaac  and
      Firat, Orhan  and
      Saxena, Ishank",
    editor = "Bouamor, Houda  and
      Pino, Juan  and
      Bali, Kalika",
    booktitle = "Proceedings of the 2023 Conference on Empirical Methods in Natural Language Processing",
    month = dec,
    year = "2023",
    address = "Singapore",
    publisher = "Association for Computational Linguistics",
    url = "https://aclanthology.org/2023.emnlp-main.26/",
    doi = "10.18653/v1/2023.emnlp-main.26",
    pages = "371--405"
}

@inproceedings{salesky-etal-2024-benchmarking,
  title = "Benchmarking Visually-Situated Translation of Text in Natural Images",
  author = "Salesky, Elizabeth  and
  Koehn, Philipp  and
  Post, Matt",
  editor = "Haddow, Barry  and
  Kocmi, Tom  and
  Koehn, Philipp  and
  Monz, Christof",
  booktitle = "Proceedings of the Ninth Conference on Machine Translation",
  month = nov,
  year = "2024",
  address = "Miami, Florida, USA",
  publisher = "Association for Computational Linguistics",
  url = "https://aclanthology.org/2024.wmt-1.115/",
  doi = "10.18653/v1/2024.wmt-1.115",
  pages = "1167--1182"
}

@inproceedings{riley-etal-2024-finding,
  title = "Finding Replicable Human Evaluations via Stable Ranking Probability",
  author = "Riley, Parker  and
  Deutsch, Daniel  and
  Foster, George  and
  Ratnakar, Viresh  and
  Dabirmoghaddam, Ali  and
  Freitag, Markus",
  editor = "Duh, Kevin  and
  Gomez, Helena  and
  Bethard, Steven",
  booktitle = "Proceedings of the 2024 Conference of the North American Chapter of the Association for Computational Linguistics: Human Language Technologies (Volume 1: Long Papers)",
  month = jun,
  year = "2024",
  address = "Mexico City, Mexico",
  publisher = "Association for Computational Linguistics",
  url = "https://aclanthology.org/2024.naacl-long.275/",
  doi = "10.18653/v1/2024.naacl-long.275",
  pages = "4908--4919"
}

@inproceedings{finkelstein-etal-2024-introducing,
    title = "Introducing the {N}ews{P}a{LM} {MBR} and {QE} Dataset: {LLM}-Generated High-Quality Parallel Data Outperforms Traditional Web-Crawled Data",
    author = "Finkelstein, Mara  and
      Vilar, David  and
      Freitag, Markus",
    editor = "Haddow, Barry  and
      Kocmi, Tom  and
      Koehn, Philipp  and
      Monz, Christof",
    booktitle = "Proceedings of the Ninth Conference on Machine Translation",
    month = nov,
    year = "2024",
    address = "Miami, Florida, USA",
    publisher = "Association for Computational Linguistics",
    url = "https://aclanthology.org/2024.wmt-1.126/",
    doi = "10.18653/v1/2024.wmt-1.126",
    pages = "1355--1372"
}
\bibliographystyle{acl_natbib}

\vfill\eject
\appendix

\section*{Contributions}

\subsection*{Core Contributors}
  Mara Finkelstein \\
  Isaac Caswell \\
  Tobias Domhan \\
  Jan-Thorsten Peter \\
  Juraj Juraska \\
  Parker Riley \\
  Daniel Deutsch \\
  Geza Kovacs{\renewcommand{\thefootnote}{$\star$}\footnote{Now at Anthropic.}}%
  \addtocounter{footnote}{-1}

\subsection*{Lead}
  David Vilar \\
  Markus Freitag \\

\subsection*{Contributors}
  Cole Dilanni \\
  Colin Cherry \\
  Eleftheria Briakou \\
  Elizabeth Nielsen \\
  Jiaming Luo \\
  Kat Black \\
  Ryan Mullins \\
  Sweta Agrawal \\
  Wenda Xu \\

\subsection*{Support}
  Erin Kats \\
  Stephane Jaskiewicz \\

\clearpage

\onecolumn
\section{Automatic metrics per language}
\label{sec:results-per-lang}
\enlargethispage{1cm}

\begin{table}[H]
  \caption{Comparison of performance of the \gemtrans~(GT) models with baseline Gemma models~(G3) for each language pair in the \wmtpp set using MetricX.}
  \label{tab:per-language-metricx}
  \begin{center}
    \scriptsize
    \rowcolors{3}{gray!20}{white} 
    \begin{tabular}{l<{\hspace{2em}}R{2}R{2}R{2}>{\hspace{3em}}lR{2}R{2}R{2}}
      \toprule
        & \multicolumn{3}{c}{\gemtrans}
        && \multicolumn{3}{c}{\gemmathree}
        \\
        \cmidrule(r){2-4} \cmidrule(l){6-8}

        & \multicolumn{1}{l}{27B}
        & \multicolumn{1}{l}{12B}
        & \multicolumn{1}{l}{4B}
        && \multicolumn{1}{l}{27B}
        & \multicolumn{1}{l}{12B}
        & \multicolumn{1}{l}{4B}
        \\
        \midrule
      en$\to$ar\_EG  & b2.5370621977529177 & 2.7770720842721253 & 3.5676041097225    && 3.316606601451834  & 3.700323820672929  & 4.601158393950512  \\
      en$\to$ar\_SA  & b2.4158107047900557 & 2.6559382762294264 & 3.4293126687717934 && 3.1883499469297627 & 3.6368640494843323 & 4.488654488542427  \\
      en$\to$bg\_BG  & b2.8037818035130235 & 3.2456590385971746 & 4.303872342555163  && 3.899498128303094  & 4.46557573126823   & 5.809210832269552  \\
      en$\to$bn\_IN  & b1.8752948303551724 & 2.1194094368101406 & 2.9220185280622295 && 2.55877143209024   & 2.8679546708008274 & 3.8612988725459823 \\
      en$\to$ca\_ES  & b3.1812426822880906 & 3.5813729065159956 & 5.204958483250811  && 4.066470370317499  & 4.88790615250667   & 6.967008726143589  \\
      en$\to$cs\_CZ  & b3.4773197118658574 & 4.032290118001401  & 5.407030595947678  && 4.6157461864718545 & 5.322496845433489  & 7.46602177649426   \\
      en$\to$da\_DK  & b2.1061878675362093 & 2.447218743075306  & 3.249719393718988  && 2.9950011487895    & 3.360654001698519  & 4.398244275913263  \\
      en$\to$de\_DE  & b1.187378672066067  & 1.357838528737193  & 1.9308940334575406 && 1.6285497400560416 & 1.9259396474148767 & 2.7203017603276143 \\
      en$\to$el\_GR  & b2.5697945914114824 & 3.340626237851878  & 4.661348156821138  && 3.7347624501135823 & 4.340276897403722  & 6.307627852990603  \\
      en$\to$es\_MX  & b1.8793455279927003 & 2.062043429914047  & 2.507930046735176  && 2.5421277300571092 & 2.7527188378502614 & 3.3532861684487822 \\
      en$\to$et\_EE  & b4.608786498654323  & 6.154765206947923  & 11.03340838458389  && 6.401561780204065  & 8.894743115548044  & 14.775356200585763 \\
      en$\to$fa\_IR  & b1.9855019547472088 & 2.2812860004448643 & 3.3350272648773776 && 2.9777804242485826 & 3.4092215857313324 & 4.767534167327297  \\
      en$\to$fi\_FI  & b3.1936586261610502 & 3.7687235592709234 & 5.682012208426992  && 4.188153235591017  & 5.111710836117466  & 7.539595539169386  \\
      en$\to$fil\_PH & b2.9801036385043216 & 3.173756803543074  & 4.200687277084216  && 3.6213657647216073 & 4.0311463823541995 & 5.221496600611135  \\
      en$\to$fr\_CA  & b2.205120938589486  & 2.370165361549395  & 2.9197661560804895 && 2.7776952935382724 & 2.971507569712897  & 3.763458040061717  \\
      en$\to$fr\_FR  & b2.188098971064513  & 2.4388154310019066 & 2.9707285227874913 && 2.7808851417774956 & 3.01437593055889   & 3.897174341014276  \\
      en$\to$gu\_IN  & b4.691508191047857  & 4.927322026155889  & 6.320239021504919  && 5.266561445345482  & 5.793036990705877  & 7.670144467428327  \\
      en$\to$he\_IL  & b2.723443709783411  & 3.123373076017015  & 4.993518272643754  && 3.9045755600585834 & 4.409905812144279  & 6.697869104752317  \\
      en$\to$hi\_IN  & b3.519957955678304  & 3.6911589987265567 & 4.33102483668675   && 4.1084652017802    & 4.357652678154409  & 5.027182878755654  \\
      en$\to$hr\_HR  & b2.0540544299253574 & 2.310591416945681  & 3.1706928837423525 && 2.6247876392832645 & 3.0841274432837964 & 4.259247298849126  \\
      en$\to$hu\_HU  & b4.243895054298142  & 5.002860390736411  & 7.838922058884054  && 5.5053690348751845 & 6.7946859192724025 & 10.746268044256915 \\
      en$\to$id\_ID  & b2.066535506758373  & 2.1706295251885117 & 2.628894549503457  && 2.721062975153715  & 2.8437153986247723 & 3.2688697188782196 \\
      en$\to$is\_IS  & b5.692071659086893  & 7.931692516990006  & 15.535486124642194 && 8.311688385810703  & 12.164855916518718 & 19.220411907322706 \\
      en$\to$it\_IT  & b1.8809301929625992 & 2.1725017773220316 & 2.636767715223444  && 2.5989383848694465 & 2.839879829576239  & 3.5985156890315313 \\
      en$\to$ja\_JP  & b3.5256539835749816 & 3.8195096885164577 & 4.439829327367867  && 4.111428534938023  & 4.300316149089485  & 5.092945293880378  \\
      en$\to$kn\_IN  & b4.177538754496102  & 4.776568905450404  & 7.109528550583248  && 5.089345000544563  & 6.819454180588946  & 10.480674909676114 \\
      en$\to$ko\_KR  & b2.8082467780448495 & 2.973006175396343  & 3.9271360421863695 && 3.425766865008821  & 3.7945444100846846 & 4.720259842618058  \\
      en$\to$lt\_LT  & b4.389305583434179  & 5.411937659567532  & 9.578815920775135  && 6.0062042052236695 & 7.708494075100559  & 13.388301802581797 \\
      en$\to$lv\_LV  & b5.688316525278302  & 7.21626311975997   & 12.117916322957415 && 7.545019038479465  & 9.895983146014624  & 15.74692487414771  \\
      en$\to$ml\_IN  & b3.635737387339274  & 4.304763685834284  & 7.331959735881537  && 4.774710508078957  & 6.83779238332063   & 11.891455817346772 \\
      en$\to$mr\_IN  & b3.165677677684774  & 3.4711459280302126 & 4.303954512315491  && 4.108708912351479  & 4.604712904865543  & 5.635049456451088  \\
      en$\to$nl\_NL  & b1.6671357286938777 & 2.0081522986467464 & 2.8413005969428924 && 2.480127318424638  & 2.8164583499000098 & 3.867121979268268  \\
      en$\to$no\_NO  & b2.0922218549453344 & 2.384863123317094  & 3.2641832169805034 && 2.9442478233211054 & 3.1665611571981573 & 4.227412300141683  \\
      en$\to$pa\_IN  & b3.6695158728398383 & 4.435136373993009  & 5.533614122308791  && 4.397866290745636  & 5.986173164658249  & 11.198660026087115 \\
      en$\to$pl\_PL  & b4.142534780222922  & 4.577208422465871  & 5.641395201689253  && 5.17383848951819   & 5.637087704318886  & 7.070160703298947  \\
      en$\to$pt\_BR  & b2.1299183102877577 & 2.3578260618805262 & 2.9282797130998612 && 2.8994847270126534 & 3.1546269779326392 & 3.7680617049656577 \\
      en$\to$pt\_PT  & b2.5511756457349595 & 2.6798292272530184 & 3.093146863410948  && 3.3862105199310464 & 3.7833929967251607 & 4.132702370853319  \\
      en$\to$ro\_RO  & b2.859971646157404  & 3.2545081137058633 & 4.182029311483105  && 3.987094971118495  & 4.393755515137067  & 5.6962282343767585 \\
      en$\to$ru\_RU  & b2.177802265573215  & 2.475623920338694  & 3.2496083309330666 && 3.012024836191752  & 3.3864323958094853 & 4.5357533904840235 \\
      en$\to$sk\_SK  & b3.8113769938548407 & 4.542497439185778  & 6.704042968774835  && 5.0429862250263495 & 5.858521793658535  & 9.168970291099201  \\
      en$\to$sl\_SI  & b3.548324222234078  & 4.24371011077892   & 7.122489970806055  && 4.5637998469406735 & 5.7328429056253905 & 9.390213777474127  \\
      en$\to$sr\_RS  & 2.775747888768092   & b2.680835775161783 & 5.657348768071582  && 3.7529252883667747 & 3.7613395702714723 & 6.939012289295594  \\
      en$\to$sv\_SE  & b1.9959424791159108 & 2.3147735770694755 & 3.1395743387169204 && 2.7277266433889356 & 3.056036389475533  & 4.1355400068374975 \\
      en$\to$sw\_KE  & b4.454249954906603  & 5.356020600348711  & 10.65162300922287  && 5.921220447154095  & 7.900122194023182  & 14.052979212844123 \\
      en$\to$sw\_TZ  & b4.302276411497345  & 5.249774988709638  & 10.301707007813578 && 5.734664589880655  & 7.8469388047854105 & 13.894021087760727 \\
      en$\to$ta\_IN  & b2.868447380885482  & 2.975050808272014  & 3.8958922981595    && 3.5330215007066728 & 3.829935298953205  & 5.038822444435209  \\
      en$\to$te\_IN  & b3.7576216453065476 & 3.9697789137562114 & 4.826784877851606  && 4.405967515396575  & 4.740005689300597  & 5.759339067588249  \\
      en$\to$th\_TH  & b2.3308502973988654 & 2.659346390484522  & 3.4879411618070058 && 2.960632913202668  & 3.190036203106865  & 4.144453081410999  \\
      en$\to$tr\_TR  & b4.177851457102224  & 4.636800565756857  & 6.168749863716463  && 5.316019427947079  & 6.017267097812146  & 8.033997758089875  \\
      en$\to$uk\_UA  & b2.975028878124431  & 3.2908606491206833 & 4.159592028971141  && 3.7938822128344327 & 4.2842621218257895 & 5.396240892556185  \\
      en$\to$ur\_PK  & b3.118360965978354  & 3.5873852419666945 & 5.6680029816304645 && 3.8557222520622116 & 4.858048198372126  & 7.80107734904935   \\
      en$\to$vi\_VN  & b1.9732366760106137 & 2.2011872163042425 & 2.86614572283191   && 2.564626371519019  & 2.878904571974029  & 3.61868765107356   \\
      en$\to$zh\_CN  & b1.8624178896852148 & 2.0685804386584397 & 2.6608424159775796 && 2.4650289935525507 & 2.605480300926138  & 3.270112274622079  \\
      en$\to$zh\_TW  & b2.0355376448307654 & 2.2119449015706776 & 2.7721922746083387 && 2.6322830101863173 & 2.8003329090677047 & 3.622934323840309  \\
      en$\to$zu\_ZA  & b6.9899714082169035 & 10.728989951188366 & 18.292104465079806 && 9.04914885389929   & 14.802177208165327 & 21.521275435450175 \\

        \midrule
        & \multicolumn{1}{l}{27B}
        & \multicolumn{1}{l}{12B}
        & \multicolumn{1}{l}{4B}
        && \multicolumn{1}{l}{27B}
        & \multicolumn{1}{l}{12B}
        & \multicolumn{1}{l}{4B}
        \\
        \hiderowcolors
        \cmidrule(r){2-4} \cmidrule(l){6-8}
        & \multicolumn{3}{c}{\gemtrans}
        && \multicolumn{3}{c}{\gemmathree} \\
        \bottomrule
      \end{tabular}
    \end{center}
  \end{table}

\section{Additional synthetic data languages}
\label{sec:synth-data-langs}

For the following languages we created synthetic data, in addition to the languages covered by \wmtpp:

English-Armenian, English-Hawaiian, English-Western Frisian, English-Corsican, English-Hmong, English-Maltese, English-Tajik, English-Samoan, English-Macedonian, English-Mongolian, English-Galician, English-Albanian, English-Uzbek, English-Uyghur, English-Belarusian, English-Sinhala, English-Basque, English-Haitian Creole, English-Bosnian, English-Kyrgyz, English-Kazakh, English-Khmer, English-Scottish Gaelic, English-Lao, English-Irish, English-Luxembourgish, English-Burmese, English-Sundanese, English-Javanese, English-Malay.

\section{Full list of languages for SFT}
\label{sec:list-of-languages}

Table~\ref{tab:sft-langs-biEnglish} shows the languages paired with English in both directions, Table~\ref{tab:sft-lang-fromEnglish} the languages paired with English as source language and Table~\ref{tab:sft-lang-nonEnglish} the languages pairs not involving English.
Together these three tables give the full language coverage of the SFT data used for \gemtrans.

\begin{table*}
  \caption{Languages paired with English in both directions.}
  \begin{tiny}
    \vspace{-0.5cm}
    \begin{center}
      \rowcolors{0}{white}{gray!20}
      \begin{tabular}{llll}
        \toprule
        Abkhaz (ab) & Acehnese (ace) & Acholi (ach) & Afar (aa)\\
        Afrikaans (af) & Ahirani (ahr) & Alur (alz) & Amharic (am)\\
        Assamese (as) & Assyrian Neo-Aramaic (aii) & Avar (av) & Awadhi (awa)\\
        Aymara (ay) & Badaga (bfq) & Bagheli (bfy) & Bagri (bgq)\\
        Balinese (ban) & Baluchi (bal) & Bambara (bm) & Banjar (Arabic script) (bjn-Arab)\\
        Banjar (bjn) & Baoul\u00e9 (bci) & Bashkir (ba) & Batak Karo (btx)\\
        Batak Simalungun (bts) & Batak Toba (bbc) & Bemba (Zambia) (bem) & Betawi (bew)\\
        Bhojpuri (bho) & Bikol (bik) & Bodo (India) (brx) & Braj (bra)\\
        Breton (br) & Buginese (bug) & Bundeli (bns) & Buryat (bua)\\
        Cantonese (yue) & Chakma (Latin script) (ccp-Latn) & Chamorro (ch) & Chechen (ce)\\
        Chhattisgarhi (hne) & Chichewa (ny) & Chinese (zh-CN) & Chittagonian (ctg)\\
        Chuukese (chk) & Chuvash (cv) & Crimean Tatar (Cyrillic script) (crh) & Crimean Tatar (Latin script) (crh-Latn)\\
        Dari (fa-AF) & Dhivehi (dv) & Dhundari (dhd) & Dinka (din)\\
        Dogri (doi) & Dombe (dov) & Dutch (nl) & Dyula (dyu)\\
        Dzongkha (dz) & East Circassian (kbd) & Eastern Huasteca Nahuatl (nhe) & Efik (efi)\\
        Egyptian Arabic (arz) & Ewe (ee) & Faroese (fo) & Fijian (fj)\\
        Fon (fon) & French (fr) & Friulian (fur) & Fulani (ff)\\
        Ga (gaa) & Garo (Latin script) (grt-Latn) & German (de) & Goan Konkani (gom)\\
        Guarani (gn) & Hakha Chin (cnh) & Hausa (ha) & Hiligaynon (hil)\\
        Hindi (hi) & Ho (Warang Chiti script) (hoc-Wara) & Hunsrik (hrx) & Iban (iba)\\
        Igbo (ig) & Ilocano (ilo) & Indonesian (id) & Inuktut (Syllabics) (iu)\\
        Isoko (iso) & Italian (it) & Jamaican Patois (jam) & Japanese (ja)\\
        Jingpo (kac) & K'iche' (quc) & Kalaallisut (kl) & Kangri (xnr)\\
        Kanuri (kr) & Kapampangan (pam) & Karakalpak (kaa) & Kashmiri (Devanagari script) (ks-Deva)\\
        Kashmiri (ks) & Kedah Malay (meo) & Khasi (kha) & Kiga (cgg)\\
        Kikuyu (ki) & Kiluba (Luba-Katanga) (lu) & Kinyarwanda (rw) & Kituba (DRC) (ktu)\\
        Kokborok (trp) & Komi (kv) & Kongo (kg) & Korean (ko)\\
        Krio (kri) & Kumaoni (kfy) & Kurdish (Sorani) (ckb) & Kurukh (kru)\\
        Lahnda Punjabi (Pakistan) (pa-Arab) & Latgalian (ltg) & Lepcha (lep) & Libyan Arabic (ayl)\\
        Ligurian (lij) & Limbu (Limbu script) (lif-Limb) & Limburgish (li) & Lingala (ln)\\
        Lombard (lmo) & Luganda (lg) & Luo (luo) & Madurese (mad)\\
        Magahi (mag) & Maithili (mai) & Makassar (mak) & Malagasy (mg)\\
        Malay (Jawi Script) (ms-Arab) & Mam (mam) & Mandeali (mjl) & Manx (gv)\\
        Mapudungun (arn) & Marshallese (mh) & Marwadi (mwr) & Mauritian Creole (mfe)\\
        Meadow Mari (chm) & Meiteilon (Manipuri) (mni-Mtei) & Mewari (mtr) & Minang (min)\\
        Mizo (lus) & Modern Standard Arabic (ar) & Moor\u00e9 (mos) & Morrocan Arabic (ar-MA)\\
        Mundari (Devanagari script) (unr-Deva) & NKo (bm-Nkoo) & Navajo (nv) & Ndau (ndc-ZW)\\
        Nepalbhasa (Newari) (new) & Nepali (ne) & Nigerian Pidgin (pcm) & Nimadi (noe)\\
        North Levantine Arabic (apc) & North Ndebele (nd) & Northern Sami (se) & Nuer (nus)\\
        Occitan (oc) & Oromo (om) & Ossetian (os) & Pangasinan (pag)\\
        Papiamento (pap) & Polish (pl) & Q'eqchi' (kek) & Quechua (qu)\\
        Rohingya (Latin script) (rhg-Latn) & Romani (rom) & Rundi (rn) & Russian (ru)\\
        Sambalpuri (spv) & Sango (sg) & Sanskrit (sa) & Santali (Latin Script) (sat-Latn)\\
        Saraiki (skr) & Sepedi (nso) & Sesotho (st) & Seychellois Creole (crs)\\
        Shan (shn) & Sherpa (Tibetan script) (xsr-Tibt) & Shina (scl) & Shona (sn)\\
        Sicilian (scn) & Silesian (szl) & Sindhi (Devanagari script) (sd-Deva) & Somali (so)\\
        South Ndebele (nr) & Spanish (es) & Sudanese Arabic (Deprecated BCP) (apd) & Surgujia (sgj)\\
        Surjapuri (sjp) & Susu (sus) & Swahili (sw) & Swati (ss)\\
        Sylheti (syl) & Tahitian (ty) & Tamazight (Latin Script) (ber-Latn) & Tamazight (Tifinagh Script) (ber)\\
        Tetum (tet) & Thai (th) & Tibetan (bo) & Tigrinya (ti)\\
        Tiv (tiv) & Tok Pisin (tpi) & Tonga (Tonga Islands) (to) & Tsonga (ts)\\
        Tswana (tn) & Tulu (tcy) & Tumbuka (tum) & Tunisian Arabic (aeb)\\
        Turkish (tr) & Tuvan (tyv) & Twi (ak) & Udmurt (udm)\\
        Venda (ve) & Venetian (vec) & Vietnamese (vi) & Wagdi (wbr)\\
        Waray (Philippines) (war) & West Circassian (ady) & Wolof (wo) & Xhosa (xh)\\
        Yakut (sah) & Yoruba (yo) & Yucatec Maya (yua) & Zapotec (zap)\\

        \bottomrule
      \end{tabular}
    \end{center}
  \end{tiny}
  \label{tab:sft-langs-biEnglish}
\end{table*}

\begin{table*}
  \caption{Languages from English}
  \begin{tiny}
    \vspace{-0.5cm}
    \begin{center}
      \rowcolors{0}{white}{gray!20}
      \begin{tabular}{lllll}
        \toprule
        Albanian (sq) & Arabic (Egypt) (ar-EG) & Armenian (hy) & Bangla (bn) & Basque (eu)\\
        Belarusian (be) & Bosnian (bs) & Bulgarian (bg) & Burmese (my) & Catalan (ca)\\
        Chinese (Taiwan) (zh-TW) & Corsican (co) & Croatian (hr) & Czech (cs) & Danish (da)\\
        Estonian (et) & Filipino (fil) & Finnish (fi) & French (Canada) (fr-CA) & Galician (gl)\\
        Greek (el) & Gujarati (gu) & Haitian Creole (ht) & Hawaiian (haw) & Hebrew (he)\\
        Hmong (hmn) & Hungarian (hu) & Icelandic (is) & Inuktut (Latin) (iu-Latn) & Irish (ga)\\
        Javanese (jv) & Kannada (kn) & Kazakh (kk) & Khmer (km) & Kyrgyz (ky)\\
        Lao (lo) & Latvian (lv) & Lithuanian (lt) & Luxembourgish (lb) & Macedonian (mk)\\
        Malay (ms) & Malayalam (ml) & Maltese (mt) & Marathi (mr) & Mongolian (mn)\\
        Norwegian (no) & Persian (fa) & Portuguese (Brazil) (pt-BR) & Portuguese (Portugal) (pt-PT) & Punjabi (pa)\\
        Romanian (ro) & Samoan (sm) & Santali (Ol Chiki script) (sat) & Scottish Gaelic (gd) & Serbian (Cyrillic) (sr-Cyrl)\\
        Serbian (Latin) (sr-Latn) & Sinhala (si) & Slovak (sk) & Slovenian (sl) & Sundanese (su)\\
        Swahili (Kenya) (sw-KE) & Swahili (Tanzania) (sw-TZ) & Swedish (sv) & Tajik (tg) & Tamil (ta)\\
        Telugu (te) & Tshiluba (Luba-Lulua) (lua) & Ukrainian (uk) & Urdu (ur) & Uyghur (ug)\\

        \bottomrule
      \end{tabular}
    \end{center}
  \end{tiny}
  \label{tab:sft-lang-fromEnglish}
\end{table*}

\begin{table*}
  \caption{Non-English language pairs}
  \begin{tiny}
    \vspace{-0.5cm}
    \begin{center}
      \rowcolors{0}{white}{gray!20}
      \begin{tabular}{lll}
        \toprule
        Amharic (am)$\leftrightarrow$Arabic (ar) & Amharic (am)$\leftrightarrow$Mandarin Chinese (zh) & Arabic (ar)$\leftrightarrow$Swahili (sw)\\
        Cantonese (yue)$\leftrightarrow$Mandarin Chinese (zh) & Cantonese (yue)$\leftrightarrow$Taiwanese Mandarin (zh-Hant) & Chinese (zh-CN)$\to$Japanese (ja)\\
        Czech (cs)$\to$German (de) & Czech (cs)$\to$Ukrainian (uk) & Mandarin Chinese (zh)$\leftrightarrow$Swahili (sw)\\

        \bottomrule
      \end{tabular}
    \end{center}
  \end{tiny}
  \label{tab:sft-lang-nonEnglish}
\end{table*}

\end{document}